\documentclass[12pt]{article}
\usepackage{graphicx}
\usepackage{setspace}
\usepackage[left=1in,right=1in,top=1in,bottom=1in]{geometry}
\usepackage{times}
\usepackage{hyperref}
\usepackage{subfigure}
\makeatletter
\renewcommand\section{\@startsection{section}{1}{\z@}%
                                  {-3.5ex \@plus -1ex \@minus -.2ex}%
                                  {2.3ex \@plus.2ex}%
                                  {\normalfont\bfseries}}
\makeatother
\begin{document}
\noindent
\footnotesize{
\noindent
\textit{Proceedings of the 6th INFORMS Workshop on Data Mining and Health Informatics (DM-HI 2011)\\
\noindent
P. Qian, Y. Zhou, C. Rudin, eds.}}

\vspace{0.1in}
\begin{center}
    {\large\bf Probabilistic Classification using Fuzzy Support Vector Machines}\\
    \vspace{0.3in}
\textbf{Marzieh Parandehgheibi}

Operations Research Center\\
Massachusetts Institute of Technology\\
Cambridge, MA\\
\texttt{marp@mit.edu}\\
\vspace{0.2in}
\end{center}

\begin{center}
    {\bf Abstract}\\
\end{center}
\noindent
In medical applications such as recognizing the type of a tumor as Malignant or Benign, a wrong diagnosis can be devastating. Methods like Fuzzy Support Vector Machines (FSVM) try to reduce the effect of misplaced training points by assigning a lower weight to the outliers. However, there are still uncertain points which are similar to both classes and assigning a class by the given information will cause errors. In this paper, we propose a two-phase classification method which probabilistically assigns the uncertain points to each of the classes. The proposed method is applied to the Breast Cancer Wisconsin (Diagnostic) Dataset which consists of 569 instances in 2 classes of Malignant and Benign. This method assigns certain instances to their appropriate classes with probability of one, and the uncertain instances to each of the classes with associated probabilities. Therefore, based on the degree of uncertainty, doctors can suggest further examinations before making the final diagnosis.


%

\section{Introduction}
Support Vector Machine (SVM) is one of the most powerful pattern classification methods known in recent years \cite{SVMtut, DataMiningBook}. In a two-class data set, SVM first maps the data set to a higher dimension to make the classes seperable. Then, it finds the seperating hyperplane while maximizing the margin between two classes. However, in many cases, the classes may have overlaps, i.e., they are not separable. In such cases, SVM allows some points to be on the wrong side with a specific cost, though it still maximizes the margin between the classes. \\

Moreover, there exist many applications, in which all points may not be exactly assigned to either of the classes, i.e., some points might be noisy. Generally, SVM assigns equal weights to all of the points. Thus, they have the same importance in determining the margin. Lin and Wang, in \cite{FSVM}, addressed this deficiency by introducing a modification of SVM called ``Fuzzy Support Vector Machine(FSVM)''. By defining a fuzzy membership function, they assigned a fuzzy value or weight for each training point, and then ran SVM to reduce the effects of outliers and noisy points. FSVM has attracted a lot of attention on both improving and applying this method to different data sets. Diffent fuzzy membership funztions have been considered to improve the performance of FSVM \cite{Jiang}, and an iterative Fuzzy Support Machine (IFSVM) method is introduced in \cite{IFSVM}. Moreover, Abe and Inoue showed that FSVM can be generalized from two-class data sets to multiclass data sets \cite{MFSVM}. \\

Another pattern classification method is fuzzy classification, which is based on the truthness of value known as \textit{membership value}. For each attribute of a class, a membership function is defined, and the range of membership values for every instance is between $[0,1]$, where $1$ shows the absolute truth and $0$ shows the absolute false. For each class, an if-then rule is defined which shows if the membership values of an instance are in the determined ranges, then the instance would be assinged to that class.\\

Despite the great performance of FSVM and fuzzy classification, there is still a shortcoming. In medical applications such as disease recognition, a wrong diagnosis can have devastating effects on a person's life. To decrease the possibility of error, we propose a two-phase classification method which probabilistically assigns the uncertain points to each of the classes. First, FSVM is applied to the whole training data such that most of the uncertain points will be placed in the margin. Moreover, the certain points are assigned to appropriate classes. Next, a fuzzy membership function and an appropriate rule are defined to classify the points that were located in the margin. This will result in assigning uncertain points to each of the classes with a specific probability. The proposed method is applied to the Breast Cancer Wisconsin (Diagnostic) Dataset which consists of 569 instances in 2 classes of Malignant and Benign. This method assigns certain instances to their appropriate classes with probability of one, and the uncertain instances to each of the classes with associated probabilities. Therefore, based on the probability values, doctors can suggest further examinations before making the final diagnosis.\\

The organization of the rest of this paper is as follows. In Section \ref{FuzzySVM}, we present the idea behind Fuzzy Support Vector Machine method. In Section \ref{Margin}, we discuss the Fuzzy classification, and propose our probabilistic method. Finally, we provide simulation results in Section \ref{SimResult} and conclusions in Section \ref{concl}.\\

\section{Fuzzy SVM}\label{FuzzySVM}

In this section, the classic SVM is explained followed by FSVM. Suppose training data consists of $N$ pairs $(x_1, y_1),\cdots,$ $(x_N, y_N)$, where $x_i \in R^p$ and $y_i \in \{-1, 1\}$. Define a hyperplane by 
\begin{equation}
{x : f(x) = x^T \beta + \beta_0 = 0},
\end{equation}
where $\beta$ is a unit vector: $\|\beta\| = 1$. A classification rule induced by $f(x)$ is 
\begin{equation}
G(x) = sign[x^T \beta + \beta_0].\\
\end{equation}

To deal with the overlap in the classes, SVM maximizes the margin between the training points for class 1 and -1 ($M$), but allows for some points to be on the wrong side of the margin. Defining the slack variables $\zeta = (\zeta_1, \zeta_2, . . . , \zeta_N)$, the constraint can be modified as follows:

\begin{equation}
y_i(x^T_i \beta_ + \beta_0) \geq M - \zeta_i, \quad \forall i=1,\cdots,N ,
\end{equation}
\begin{equation}
\sum_{i=1}^N{\zeta_i} \leq \mbox{constant}, \quad \zeta_i \geq 0 \quad \forall i=1,\cdots,N.
\end{equation}

Defining $M = \frac{1}{\|\beta\|}$, we will have :
\begin{equation}
\textrm{min} \quad\quad \|\beta\| + \sum_{i=1}^N{\zeta_i}
\end{equation}

\begin{equation}
\textrm{s.t.} \quad\quad y_i(x^T_i \beta_ + \beta_0) \geq 1 - \zeta_i, \quad \forall i=1,\cdots,N
\end{equation}

\begin{equation}
\zeta_i \geq 0, \quad \forall i=1,\cdots,N\\
\end{equation}

As explained earlier, in many real instances, all the data points do not have the same certaintity. Therefore, the uncertain points should get a lower weight, and have less contribution in determining the marginal region. In this paper, we consider a Gaussian function for the weights.\\

Suppose out of $N$ points, $N_1$ points are in class 1 and $N_2$ remaining points are in class 2. Define the weight for each point as following:
\begin{equation}
W(x_i)= \prod_{j=1}^p exp^{\frac{-(x_{ij}-\mu_{jk})^2}{2\sigma_{jk}^2}},\quad \forall x_i \in \mbox{Class} \ k,
\end{equation}

where $\mu_{jk}$ and $\sigma_{jk}$ refer to the mean and standard deviation of $j^{th}$ feature of all points in the class $k$, respectively. Moreover, $x_{ij}$ indicates the $j^{th}$ feature value of $i^{th}$ point.\\

Then, we normalize the weights such that the total sum of the weights is equal to $N$, which is the sum of error costs for the classic SVM. In (\ref{norm}), $W_n(x_i)$ indicates the normalized weight.
\begin{equation}\label{norm}
W_n(x_i)= \frac{N}{\sum_{i=1}^N W(x_i)}W(x_i).\\
\end{equation}

Finally, the wights show up in the objective function:
\begin{equation}
min \quad\quad \|\beta\| + \sum_{i=1}^N{W_n(x_i)\zeta_i}.\\
\end{equation}

Points near to the center of each class have a higher weight than those farer. Therefore, near points will be classified certainly, and the points which are in the middle of the two classes, called uncertain points, will be located in the margin. In the next section, we discuss how to classify the marginal points probabilistically.\\

\section{Fuzzy Classification of Marginal Points}\label{Margin}
In this part, we apply a fuzzy classification on the marginal points. Here, we use a fuzzy rule-based classification method, which has been applied to many data sets \cite{Ishi19991}, \cite{Ishi19992}. The method used to generate the fuzzy rule is based on the mean and the standard deviation of each attribute \cite{Ravi2003}. Similar to the Gaussian weights in the previous section, a Gaussian fuzzy membership function $A_{ik}$ is defined for every test point $y_i$ located in the margin as 

\begin{equation}
A_{ik}= \prod_{j=1}^p exp^{\frac{-(y_{ij}-\mu_{jk})^2}{2\sigma_{jk}^2}},\quad \forall k \in {1,2},
\end{equation}

where $\mu_{jk}$ and $\sigma_{jk}$ are the mean and standard deviation of training points of class $k$ located in the margin, respectively. This membersip shows the closeness of element $y_i$ to the center of $k^{th}$ class. To measure the related closeness of a point to both centers, a ``membership probability'' is defined for each marginal point as follows:

\begin{equation}
P_{i,C1}=\frac{A_{i,C1}}{A_{i,C1}+A_{i,C2}} \quad \mbox{and} \quad P_{i,C2}=1-P_{i,C1}.\\
\end{equation}

Points with probability more than $90\%$ in class, will be assigned to that class. Otherwise, the given information is not sufficient to make a decision. Applying this probabilistic FSVM (PFSVM) on the Breast Cancer Wisconsin (Diagnostic) Dataset consisting of 569 instances, we will show that the probability of making a wrong decision is less than $1.23\%$, and the wrongly classified points in the FSVM will be determined as unknown classes, and will need additional information.\\

\section{Simulation Results}\label{SimResult}

In this section, we examine the performance of PFSVM, and compare it with previous known methods. The data set we use is the Wisconsin breast cancer diagnostic dataset \cite{wisconsin}, which consists of 569 instances in two classes of Malignan (M) and Benign (B) with 32 features per instance. First, the number of features is reduced from 32 to 23 by saving just one feature out of every set of features with correlation more than 0.95. Then by a 10-fold cross validation method the set of training and test data is determined. The two-phase probabilistic classifier is applied to the the new dataset.\\

\begin{figure}[!h]
\centering
\subfigure[SVM - Width of Margin: 0.895]
{\label{fig:SVM}\includegraphics[scale=0.07]{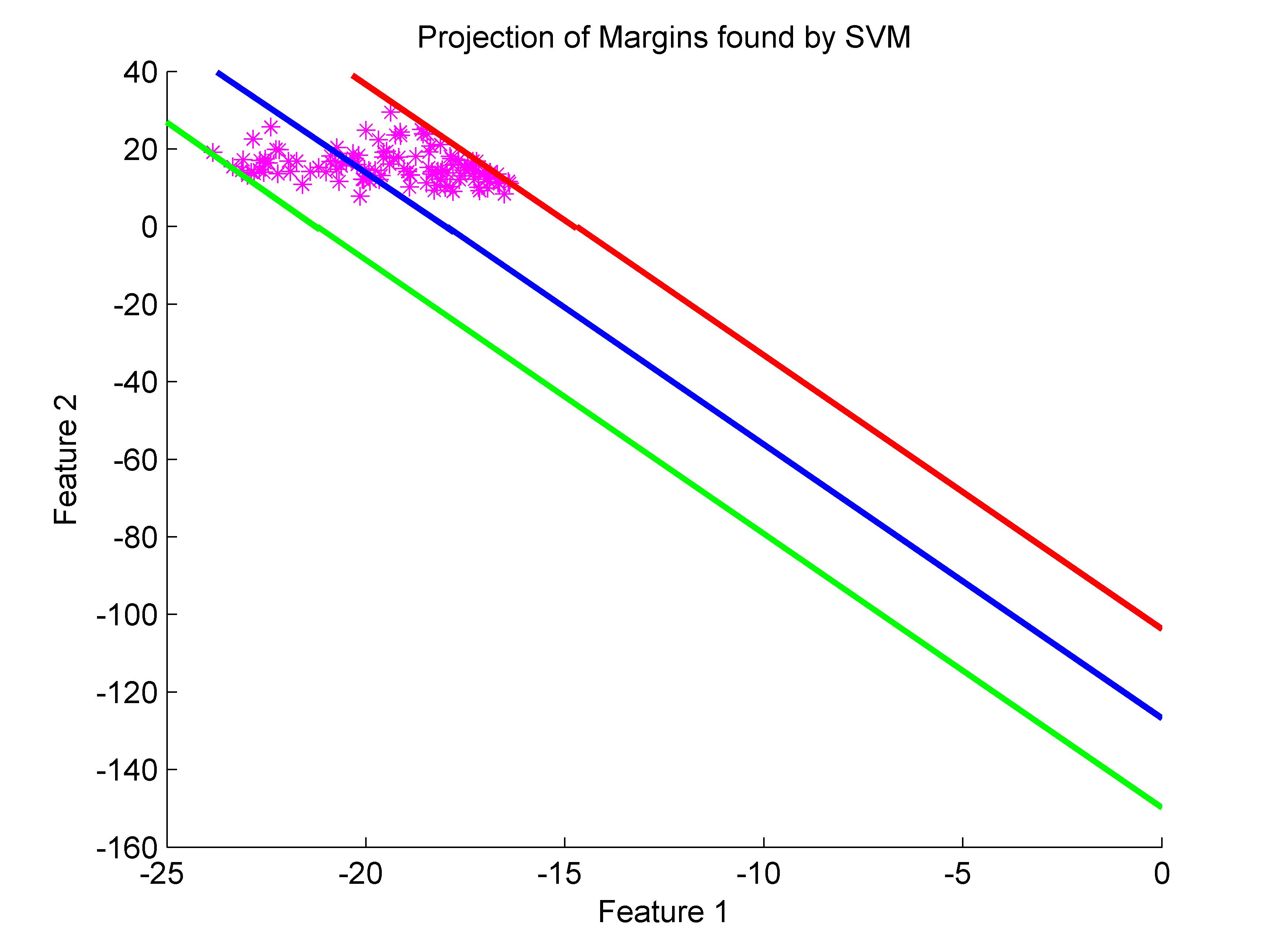}}   
\subfigure[FSVM - Width of Margin: 1.931]
{\label{fig:FSVM}\includegraphics[scale=0.7]{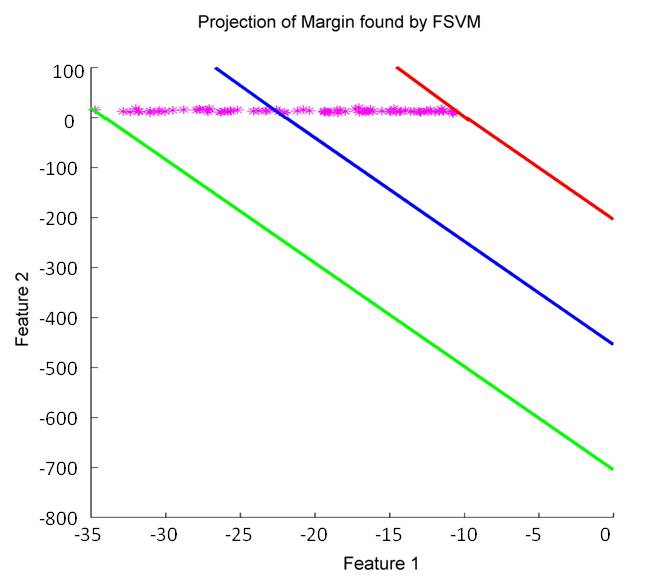}}
\caption{Comparison of margins in SVM and FSVM}
\label{widenmargin}
\end{figure}

In the first phase, we have obtained the margins using FSVM in which each training point gets a Gaussian weight. We then find the training points that are located inside the margin. Since there are 23 features, we project both margin and the points inside it onto a plane. Figure \ref{widenmargin} compares the size of margins obtained by SVM and FSVM methods.

\begin{figure}[!h]
\centering
\includegraphics[scale=0.1]{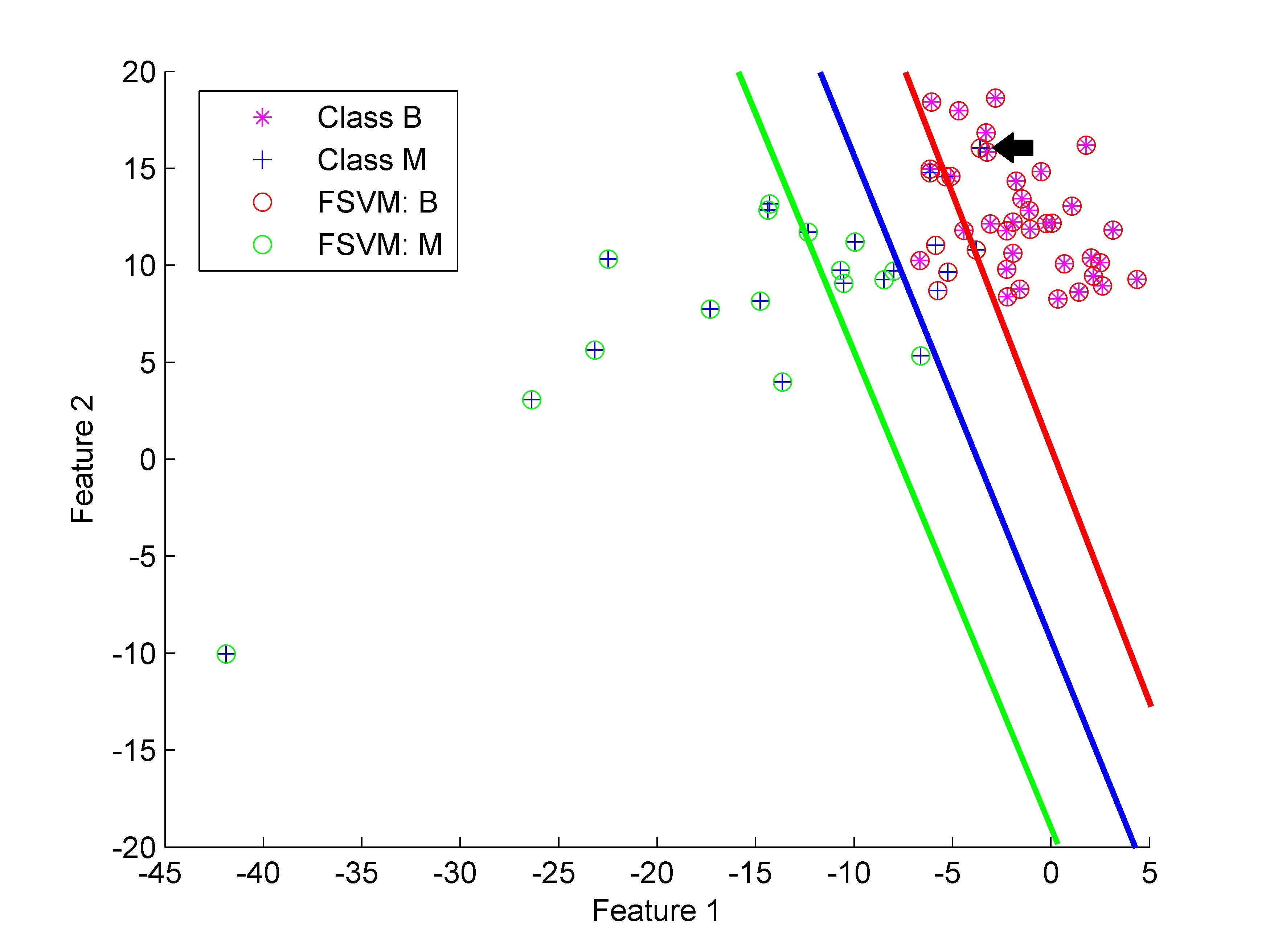}   
\caption{Errors located inside and outside of margin}
\label{errorinmargin}
\end{figure}

Moreover, observe that, on average, more than 80\% of errors are located in the margin. Figure \ref{errorinmargin} illustrates an example. We can also observe that in all the cases, a Manignan cancer is classified in the Benign group which is more dangerous than misclassifying a benign type. Therefore, we double the weight of an error of type Manign.

By doubling the cost of misclassification of an ``M'' type instance, the errors occur in the margin with probability more than 98\%. For comparison, we obtain the classifiers using both SVM and Fuzzy methods , and then apply it onto the test dataset. We also apply the probabilistic method explained in section \ref{Margin} to the whole dataset, instead of marginal points found by FSVM. Table \ref{comp} shows the comparison.

\begin{table}[!h]
\centering
\begin{tabular}{|l|c|c|c|c|c|c|c|c|c|c|c|}
  \hline
  Method \textbackslash Run\ \# & 1 & 2 & 3 & 4 & 5 & 6 & 7 & 8 & 9 & 10 & $percent_{ave}$\\ \hline
  $SVM_{err}$ & 1 & 1 & 5 & 3 & 4 & 1 & 2 & 4 & 0 & 1 & 3.86\\   \hline
  $FSVM_{err}$ & 4 & 4 & 5 & 7 & 7 & 3 & 2 & 1 & 3 & 4 & 7.02\\   \hline
  $FUZZY_{err}$ & 3 & 3 & 5 & 8 & 4 & 6 & 5 & 3 & 3 & 1 & 7.19\\   \hline
  $PFSVM_{err}$ & 1+1 & 1 & 0 & 0+2 & 3+1 & 0+2 & 0 & 1 & 2+1 & 0+2 & 1.63 
  \\   \hline
  $PFSVM_{undet}$ & 1 & 1 & 1 & 2 & 0 & 1 & 2 & 1 & 0 & 0 & 1.58\\   \hline
\end{tabular}
\caption{Comparison of different classification methods}
\label{comp}
\end{table}

Table \ref{comp} shows that the error of probabilistic method is significantly smaller than the other methods. Note that the error of FSVM is higher than SVM which is due to the fact that the decreament of points' weights allows more errors and increases the margin. However, this increament of margin ensures us that the points outside of the margin are classified correctly.

\begin{table}[!h]
\centering
\begin{tabular}{|l|c|c|c|c|c|c|c|c|c|c|c|}
  \hline
  Method \textbackslash Run\ \# & 1 & 2 & 3 & 4 & 5 & 6 & 7 & 8 & 9 & 10 & $percent_{ave}$\\ \hline
  $SVM_{err}$ & 4 & 2 & 4 & 2 & 1 & 2 & 3 & 1 & 2 & 3 & 4.29\\   \hline
  $FSVM_{err}$ & 6 & 2 & 3 & 2 & 1 & 2 & 4 & 1 & 4 & 5 & 5.36\\   \hline
  $FUZZY_{err}$ & 7 & 2 & 5 & 4 & 1 & 3 & 6 & 3 & 4 & 4 & 6.96\\   \hline
  $PFSVM_{err}$ & 0 & 2 & 1 & 0 & 0 & 2 & 0 & 0 & 0+1 & 1 & 1.23 
  \\   \hline
  $PFSVM_{undet}$ & 1 & 0 & 3 & 1 & 0 & 1 & 1 & 1 & 3 & 1 & 2.14\\   \hline
\end{tabular}
\caption{Comparison of different classification methods - double cost}
\label{comp2}
\end{table}

Table \ref{comp2}, shows the result of doubling the cost of errors in diagnosing Malignant cancer, i.e. if a cancer is Malignant, and diagnosed as Benign, this error has a higher cost than the reverse error. We can see that the classification rate is higher than 98.77\% (error is 1.23\%). Moreover, the sum of \textit{error} and \textit{undetermined} percentages is 3.37, which is less than all errors in both tables. Note that by running a probabilistic classification on the pure SVM, FSVM or Fuzzy classification, the sum of \textit{error} and \textit{undetermined} percentages will not be smaller than the current error. Therefore, neither of these classification methods can do better than the new probabilistic method.


\section{Conclusion}\label{concl}

In this paper, we considered the problem of disease diagnosis. We showed that in some cases, a deterministic classification method is not a proper method since the information may not be enough for making decision and mistakes can be devastating. Therefore, we came up with a probabilistic method and showed that it gives a smaller error rate than classic SVM and fuzzy classification. By applying our method to Breast Cancer Wisconsin (Diagnostic) Dataset, we showed that this method assigns a certain class to the cases that have strong evidence of being a Malignant or Benign type of cancer, where it assigns a probability to the cases which do not have enough information of being in a certain type.

\bibliographystyle{plain}

\end{document}